\documentclass{article}
\usepackage[a4paper, total={6in, 9in}]{geometry}

\usepackage{hyperref}
\usepackage{url}
\usepackage{booktabs}
\usepackage{amsfonts}
\usepackage{nicefrac}
\usepackage{microtype}
\usepackage{color}
\usepackage[]{graphicx}
\usepackage{amsmath}
\usepackage{xcolor}
\usepackage{paralist}
\usepackage{enumerate}
\usepackage[]{graphicx}
\usepackage[]{subcaption}
\usepackage{hyperref}

\usepackage{titling}
\usepackage{authblk}

\usepackage{threeparttable}
\usepackage{array}
\newcolumntype{C}[1]{>{\centering\let\newline\\\arraybackslash\hspace{0pt}}m{#1}}

\usepackage{caption} 
\newcommand{\draft}[1]{}

\pretitle{\begin{center} \LARGE \bf}
\posttitle{\par\end{center}\vskip 0.5em}
\preauthor{
    \begin{center}
    \normalsize \lineskip 0.5em%
    \begin{tabular}[t]{c}
}
\postauthor{\end{tabular}\par\end{center}}

\predate{\begin{center}\large}
\postdate{\par\end{center}}

\title{A simple neural network module\\for relational reasoning}

\author{Adam Santoro\thanks{Equal contribution.}}
\author{David Raposo\protect\footnotemark[1]}
\author{David G.T. Barrett}
\author{Mateusz Malinowski}
\author{\authorcr Razvan Pascanu}
\author{Peter Battaglia}
\author{Timothy Lillicrap}
\affil{
    \texttt{\normalsize adamsantoro@, draposo@, barrettdavid@, mateuszm@,} \\
    \texttt{\normalsize razp@, peterbattaglia@, countzero@google.com}
}
\affil{DeepMind\\London, United Kingdom}

\date{}

\begin{document}
\maketitle

\begin{abstract}
Relational reasoning is a central component of generally intelligent behavior, but has proven difficult for neural networks to learn. In this paper we describe how to use Relation Networks (RNs) as a simple plug-and-play module to solve problems that fundamentally hinge on relational reasoning. We tested RN-augmented networks on three tasks: visual question answering using a challenging dataset called CLEVR, on which we achieve state-of-the-art, super-human performance; text-based question answering using the bAbI suite of tasks; and complex reasoning about dynamic physical systems. Then, using a curated dataset called Sort-of-CLEVR we show that powerful convolutional networks do not have a general capacity to solve relational questions, but can gain this capacity when augmented with RNs. Our work shows how a deep learning architecture equipped with an RN module can implicitly discover and learn to reason about entities and their relations.
\end{abstract}

\section{Introduction}
\begin{figure}
    \centering
    \includegraphics[width=0.75\textwidth]{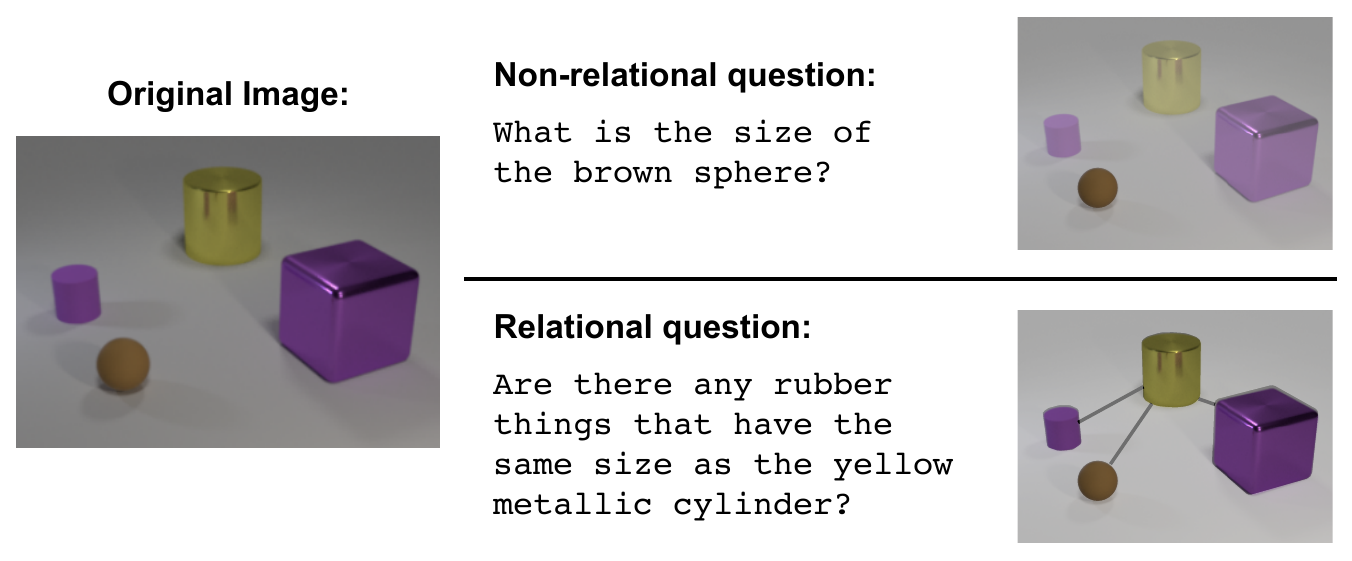}
    \caption{\textbf{An illustrative example from the CLEVR dataset of relational reasoning}. An image containing four objects is shown alongside non-relational and relational questions. The relational question requires explicit reasoning about the relations between the four objects in the image, whereas the non-relational question requires reasoning about the attributes of a particular object.}
    \label{fig:relational_reasoning}
\end{figure}

The ability to reason about the relations between entities and their properties is central to generally intelligent behavior (Figure \ref{fig:relational_reasoning}) \cite{kemp2008discovery,johnson2016clevr}. Consider a child proposing a race between the two trees in the park that are furthest apart: the pairwise distances between every tree in the park must be inferred and compared to know where to run. Or, consider a reader piecing together evidence to predict the culprit in a murder-mystery novel: each clue must be considered in its broader context to build a plausible narrative and solve the mystery.

Symbolic approaches to artificial intelligence are inherently relational \cite{newell1980physical,harnad1990symbol}. Practitioners define the relations between symbols using the language of logic and mathematics, and then reason about these relations using a multitude of powerful methods, including deduction, arithmetic, and algebra. But symbolic approaches suffer from the symbol grounding problem and are not robust to small task and input variations \cite{harnad1990symbol}. Other approaches, such as those based on statistical learning, build representations from raw data and often generalize across diverse and noisy conditions \cite{lecun2015deep}. However, a number of these approaches, such as deep learning, often struggle in data-poor problems where the underlying structure is characterized by sparse but complex relations \cite{garnelo2016towards,lake2016building}. Our results corroborate these claims, and further demonstrate that seemingly simple relational inferences are remarkably difficult for powerful neural network architectures such as convolutional neural networks (CNNs) and multi-layer perceptrons (MLPs). 

Here, we explore ``Relation Networks'' (RN) as a general solution to relational reasoning in neural networks. RNs are architectures whose computations focus explicitly on relational reasoning \cite{raposo2017discovering}. Although several other models supporting relation-centric computation have been proposed, such as Graph Neural Networks, Gated Graph Sequence Neural Networks, and Interaction Networks, \cite{scarselli2009graph,li2015gated,battaglia2016interaction}, RNs are simple, plug-and-play, and are exclusively focused on flexible relational reasoning. Moreover, through joint training RNs can influence and shape upstream representations in CNNs and LSTMs to produce implicit object-like representations that it can exploit for relational reasoning. We applied an RN-augmented architecture to CLEVR \cite{johnson2016clevr}, a recent visual question answering (QA) dataset on which state-of-the-art approaches have struggled due to the demand for rich relational reasoning. Our networks vastly outperformed the best generally-applicable visual QA architectures, and achieve state-of-the-art, super-human performance. RNs also solve CLEVR from state descriptions, highlighting their versatility in regards to the form of their input. We also applied an RN-based architecture to the bAbI text-based QA suite \cite{weston2015towards} and solved 18/20 of the subtasks. Finally, we trained an RN to make challenging relational inferences about complex physical systems and motion capture data. The success of RNs across this set of substantially dissimilar task domains is testament to the general utility of RNs for solving problems that require relation reasoning.

\section{Relation Networks}
\label{RN_architecture}

An RN is a neural network module with a structure primed for relational reasoning. The design philosophy behind RNs is to constrain the functional form of a neural network so that it captures the core common properties of relational reasoning. In other words, the capacity to compute relations is baked into the RN architecture without needing to be learned, just as the capacity to reason about spatial, translation invariant properties is built-in to CNNs, and the capacity to reason about sequential dependencies is built into recurrent neural networks.

In its simplest form the RN is a composite function:
\begin{align} 
    \text{RN}(O) = f_{\phi}\left(\sum_{i,j}g_{\theta}(o_i, o_j)\right),
    \label{eq:RN}
\end{align} 
where the input is a set of ``objects'' $O = \{o_1, o_2, ..., o_n \}$, $o_i \in \mathbb{R}^m$ is the $i^{{th}}$ object, and $f_\phi$ and $g_\theta$ are functions with parameters $\phi$ and $\theta$, respectively. For our purposes, $f_\phi$ and $g_\theta$ are MLPs, and the parameters are learnable synaptic weights, making RNs end-to-end differentiable. We call the output of $g_\theta$ a ``relation''; therefore, the role of $g_\theta$ is to infer the ways in which two objects are related, or if they are even related at all. 

RNs have three notable strengths: they learn to infer relations, they are data efficient, and they  operate on a set of objects -- a particularly general and versatile input format -- in a manner that is order invariant. 

\paragraph{RNs learn to infer relations} The functional form in Equation \ref{eq:RN} dictates that an RN should consider the potential relations between \emph{all} object pairs. This implies that an RN is not necessarily privy to which object relations actually exist, nor to the actual meaning of any particular relation. Thus, RNs must \emph{learn to infer} the existence and implications of object relations. 

In graph theory parlance, the input can be thought of as a complete and directed graph whose nodes are objects and whose edges denote the object pairs whose relations should be considered. Although we focus on this ``all-to-all'' version of the RN throughout this paper, this RN definition can be adjusted to consider only some object pairs. Similar to Interaction Networks \cite{battaglia2016interaction}, to which RNs are related, RNs can take as input a list of only those pairs that should be considered, if this information is available. This information could be explicit in the input data, or could perhaps be extracted by some upstream mechanism. 

\paragraph{RNs are data efficient} RNs use a single function $g_{\theta}$ to compute each relation. This can be thought of as a single function operating on a batch of object pairs, where each member of the batch is a particular object-object pair from the same object set. This mode of operation encourages greater generalization for computing relations, since $g_{\theta}$ is encouraged not to over-fit to the features of any particular object pair. Consider how an MLP would learn the same function. An MLP would receive \emph{all} objects from the object set simultaneously as its input. It must then learn and embed $n^2$ (where $n$ is the number of objects) identical functions within its weight parameters to account for all possible object pairings. This quickly becomes intractable as the number of objects grows. Therefore, the cost of learning a relation function $n^2$ times using a single feedforward pass per sample, as in an MLP, is replaced by the cost of $n^2$ feedforward passes per object set (i.e., for each possible object pair in the set) and learning a relation function just once, as in an RN.

\paragraph{RNs operate on a set of objects} The summation in Equation \ref{eq:RN} ensures that the RN is invariant to the order of objects in the input. This invariance ensures that the RN's input respects the property that sets are order invariant, and it ensures that the output is order invariant. Ultimately, this invariance ensures that the RN's output contains information that is generally representative of the relations that exist in the object set.

\section{Tasks}
We applied RN-augmented networks to a variety of tasks that hinge on relational reasoning. To demonstrate the versatility of these networks we chose tasks from a number of different domains, including visual QA, text-based QA, and dynamic physical systems.

\subsection{CLEVR}
In visual QA a model must learn to answer questions about an image (Figure \ref{fig:relational_reasoning}). This is a challenging problem domain because it requires high-level scene understanding \cite{antol2015vqa,malinowski14nips}. Architectures must perform complex relational reasoning -- spatial and otherwise -- over the features in the visual inputs, language inputs, and their conjunction. However, the majority of visual QA datasets require reasoning in the absence of fully specified word vocabularies, and perhaps more perniciously, a vast and complicated  knowledge of the world that is not available in the training data. They also contain ambiguities and exhibit strong linguistic biases that allow a model to learn answering strategies that exploit those biases, without reasoning about the visual input \cite{antol2015vqa,malinowski2016ask, ren2015image}. 

To control for these issues, and to distill the core challenges of visual QA, the CLEVR visual QA dataset was developed \cite{johnson2016clevr}. CLEVR contains images of 3D-rendered objects, such as spheres and cylinders (Figure \ref{fig:architecture}). Each image is associated with a number of questions that fall into different categories. For example, \texttt{query attribute} questions may ask ``\emph{What is the color of the sphere?}'', while \texttt{compare attribute} questions may ask ``\emph{Is the cube the same material as the cylinder?}''. 

For our purposes, an important feature of CLEVR is that many questions are explicitly relational in nature. Remarkably, powerful QA architectures \cite{yang2016stacked} are unable to solve CLEVR, presumably because they cannot handle core relational aspects of the task. For example, as reported in the original paper a model comprised of ResNet-101 image embeddings with LSTM question processing and augmented with stacked attention modules vastly outperformed other models at an overall performance of $68.5\%$ (compared to $52.3\%$ for the next best, and $92.6\%$ human performance) \cite{johnson2016clevr}. However, for \texttt{compare attribute} and \texttt{count} questions (i.e., questions heavily involving relations across objects), the model performed little better than the simplest baseline, which answered questions solely based on the probability of answers in the training set for a given question category (Q-type baseline).

We used two versions of the CLEVR dataset: (i) the pixel version, in which images were represented in standard 2D pixel form, and (ii) a state description version, in which images were explicitly represented by state description matrices containing factored object descriptions. Each row in the matrix contained the features of a single object -- 3D coordinates (x, y, z); color (r, g, b); shape (cube, cylinder, etc.); material (rubber, metal, etc.); size (small, large, etc.). When we trained our models, we used \emph{either} the pixel version or the state description version, depending on the experiment, but not both together.

\subsection{Sort-of-CLEVR}
\label{sec:sort_of_clevr}

To explore our hypothesis that the RN architecture is better suited to general relational reasoning as compared to more standard neural architectures, we constructed a dataset similar to CLEVR that we call ``Sort-of-CLEVR''\footnote{The ``Sort-of-CLEVR'' dataset will be made publicly available online.}. This dataset separates relational and non-relational questions.

Sort-of-CLEVR consists of images of 2D colored shapes along with questions and answers about the images. Each image has a total of $6$ objects, where each object is a randomly chosen shape (square or circle). We used $6$ colors (red, blue, green, orange, yellow, gray) to unambiguously identify each object. Questions are hard-coded as fixed-length binary strings to reduce the difficulty involved with natural language question-word processing, and thereby remove any confounding difficulty with language parsing. For each image we generated 10 relational questions and 10 non-relational questions. Examples of relational questions are: ``\emph{What is the shape of the object that is farthest from the gray object?}''; and ``\emph{How many objects have the same shape as the green object?}''.  Examples of non-relational questions are: ``What is the shape of the \textit{gray} object?''; and ``\emph{Is the blue object on the top or bottom of the scene?}''. The dataset is also visually simple, reducing complexities involved in image processing. 

\subsection{bAbI}
bAbI is a pure text-based QA dataset \cite{weston2015towards}. There are $20$ tasks, each corresponding to a particular type of reasoning, such as deduction, induction, or counting. Each question is associated with a set of supporting facts. For example, the facts ``\emph{Sandra picked up the football}'' and  ``\emph{Sandra went to the office}'' support the question ``\emph{Where is the football?}'' (answer: ``\emph{office}''). A model succeeds on a task if its performance surpasses $95\%$. Many memory-augmented neural networks have reported impressive results on bAbI. When training jointly on all tasks using $10K$ examples per task, Memory Networks pass $14/20$, DNC $18/20$, Sparse DNC $19/20$, and EntNet $16/20$ (the authors of EntNets report state-of-the-art at $20/20$; however, unlike previously reported results this was not done with joint training on all tasks, where they instead achieve $16/20$) \cite{weston2014memory, graves2016hybrid, rae2016scaling, henaff2016tracking}. 

\subsection{Dynamic physical systems}
We developed a dataset of simulated physical mass-spring systems using the MuJoCo physics engine \cite{todorov2012mujoco}. Each scene contained $10$ colored balls moving on a table-top surface. Some of the balls moved independently, free to collide with other balls and the barrier walls. Other randomly selected ball pairs were connected by invisible springs or a rigid constraint. These connections prevented the balls from moving independently, due to the force imposed through the connections. Input data consisted of state descriptions matrices, where each ball was represented as a row in a matrix with features representing the RGB color values of each object and their spatial coordinates ($x, y$) across $16$ sequential time steps.

The introduction of random links between balls created an evolving physical system with a variable number ``systems'' of connected balls (where ``systems'' refers to connected graphs with balls as nodes and connections between balls as edges). We defined two separate tasks: 1) infer the existence or absence of connections between balls when only observing their color and coordinate positions across multiple sequential frames, and 2) count the number of systems on the table-top, again when only observing each ball's color and coordinate position across multiple sequential frames. 

Both of these tasks involve reasoning about the relative positions and velocities of the balls to infer whether they are moving independently, or whether their movement is somehow dependent on the movement of other balls through invisible connections. For example, if the distance between two balls remains similar across frames, then it can be inferred that there is a connection between them. The first task makes these inferences explicit, while the second task demands that this reasoning occur implicitly, which is much more difficult. For further information on all tasks, including videos of the dynamic systems, see the supplementary information.

\section{Models}
In their simplest form RNs operate on \emph{objects}, and hence do not explicitly operate on images or natural language. A central contribution of this work is to demonstrate the flexibility with which relatively unstructured inputs, such as CNN or LSTM embeddings, can be considered as a set of objects for an RN. Although the RN expects object representations as input, the semantics of what an object \emph{is} need not be specified. Our results below demonstrate that the learning process induces upstream processing, comprised of conventional neural network modules, to produce a set of useful ``objects'' from distributed representations. 

\paragraph{Dealing with pixels}
\label{QA_RN_architecture}

\begin{figure}[h!]
	\centering
	\includegraphics[width=0.95\textwidth]{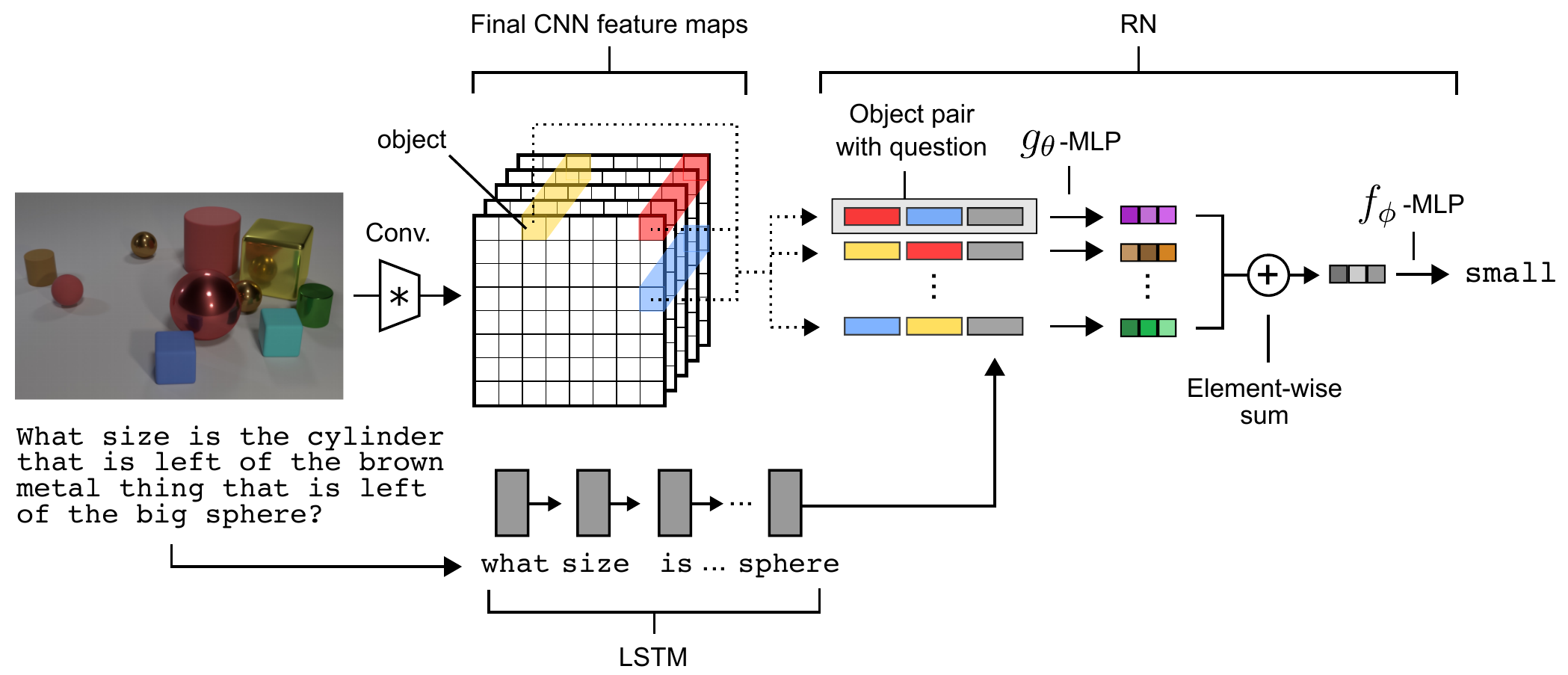}
    \caption{\textbf{Visual QA architecture}. Questions are processed with an LSTM to produce a question embedding, and images are processed with a CNN to produce a set of objects for the RN. Objects (three examples illustrated here in yellow, red, and blue) are constructed using feature-map vectors from the convolved image. The RN considers relations across all pairs of objects, conditioned on the question embedding, and integrates all these relations to answer the question.}
    \label{fig:architecture}
\end{figure}

We used a CNN to parse pixel inputs into a set of objects. The CNN took images of size $128\times128$ and convolved them through four convolutional layers to $k$ feature maps of size $d \times d$, where $k$ is the number of kernels in the final convolutional layer. We remained agnostic as to what particular image features should constitute an object. So, after convolving the image, each of the $d^2$ $k$-dimensional cells in the $d \times d$ feature maps was tagged with an arbitrary coordinate indicating its relative spatial position, and was treated as an object for the RN (see Figure \ref{fig:architecture}). This means that an ``object'' could comprise the background, a particular physical object, a texture, conjunctions of physical objects, etc., which affords the model great flexibility in the learning process. 

\paragraph{Conditioning RNs with question embeddings} The existence and meaning of an object-object relation should be question dependent. For example, if a question asks about a large sphere, then the relations between small cubes are probably irrelevant. So, we modified the RN architecture such that $g_{\theta}$ could condition its processing on the question: $a = f_{\phi}(\sum_{i,j}g_{\theta}(o_i, o_j, q))$. To get the question embedding $q$, we used the final state of an LSTM that processed question words. Question words were assigned unique integers, which were then used to index a learnable lookup table that provided embeddings to the LSTM. At each time-step, the LSTM received a single word embedding as input, according to the syntax of the English-encoded question. 

\paragraph{Dealing with state descriptions}
\label{simple_QA_RN_architecture}
We can provide state descriptions directly into the RN, since state descriptions are pre-factored object representations. Question processing can proceed as before: questions pass through an LSTM using a learnable lookup embedding for individual words, and the final state of the LSTM is concatenated to each object-pair.

\paragraph{Dealing with natural language}
\label{text_QA_RN_architecture}
For the bAbI suite of tasks the natural language inputs must be transformed into a set of objects. This is a distinctly different requirement from visual QA, where objects were defined as spatially distinct regions in convolved feature maps. So, we first identified up to $20$ sentences in the support set that were immediately prior to the probe question. Then, we tagged these sentences with labels indicating their relative position in the support set, and processed each sentence word-by-word with an LSTM (with the same LSTM acting on each sentence independently). We note that this setup invokes minimal prior knowledge, in that we delineate objects as sentences, whereas previous bAbI models processed all word tokens from all support sentences sequentially. It's unclear how much of an advantage this prior knowledge provides, since period punctuation also unambiguously delineates sentences for the token-by-token processing models. The final state of the sentence-processing-LSTM is considered to be an object. Similar to visual QA, a separate LSTM produced a question embedding, which was appened to each object pair as input to the RN. Our model was trained on the joint version of bAbI (all $20$ tasks simultaneously), using the full dataset of $10K$ examples per task. 

\paragraph{Model configuration details}
For the CLEVR-from-pixels task we used: $4$ convolutional layers each with $24$ kernels, ReLU non-linearities, and batch normalization; $128$ unit LSTM for question processing; $32$ unit word-lookup embeddings; four-layer MLP consisting of $256$ units per layer with ReLU non-linearities for $g_{\theta}$; and a three-layer MLP consisting of $256$, $256$ (with $50\%$ dropout), and $29$ units with ReLU non-linearities for $f_{\phi}$. The final layer was a linear layer that produced logits for a softmax over the answer vocabulary. The softmax output was optimized with a cross-entropy loss function using the Adam optimizer with a learning rate of $2.5e^{-4}$. We used size $64$ mini-batches and distributed training with $10$ workers synchronously updating a central parameter server. The configurations for the other tasks are similar, and can be found in the supplementary information.

We'd like to emphasize the simplicity of our overall model architecture compared to the visual QA architectures used on CLEVR thus far, which use ResNet or VGG embeddings, sometimes with fine-tuning, very large LSTMs for language encoding, and further processing modules, such as stacked or iterative attention, or large fully connected layers (upwards of 4000 units, often) \cite{johnson2016clevr}.

\section{Results}
\subsection{CLEVR from pixels}

\begin{figure}[h]
    \centering
    \includegraphics[width=0.90\textwidth]{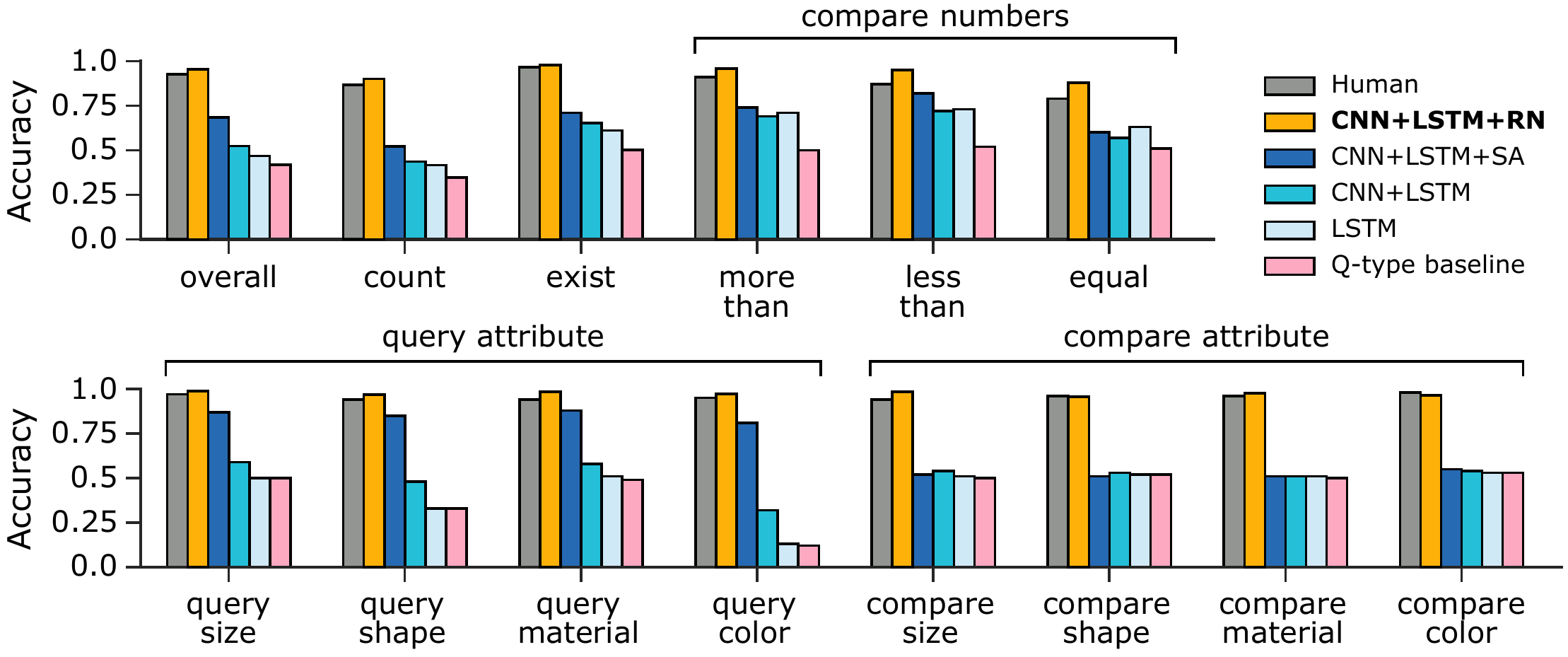}
    \caption{\textbf{Results on CLEVR from pixels.} The RN augmented model outperformed all other models and exhibited super-human performance overall. In particular, it solved ``compare attribute'' questions, which trouble all other models because they heavily depend on relational reasoning.}
   	\label{fig:bar_plot}
\end{figure}

Our model achieved state-of-the-art performance on CLEVR at $95.5\%$, exceeding the best model trained only on the pixel images and questions at the time of the dataset's publication by $27\%$, and surpassing human performance in the task (see Table \ref{table:main} and Figure \ref{fig:bar_plot}).

These results -- in particular, those obtained in the \texttt{compare attribute} and \texttt{count} categories  -- are a testament to the ability of our model to do relational reasoning. In fact, it is in these categories that state-of-the-art models struggle most. Furthermore, the relative simplicity of the network components used in our model suggests that the difficulty of the CLEVR task lies in its relational reasoning demands, not on the language or the visual processing. 

\paragraph{Results using privileged training information}A more recent study reports overall performance of 96.9\% on CLEVR, but uses additional supervisory signals on the functional programs used to generate the CLEVR questions \cite{johnson2017infer}. It is not possible for us to directly compare this to our work since we do not use these additional supervision signals. Nonetheless, our approach greatly outperforms a version of their model that was not trained with these extra signals, and even versions of their model trained using $9K$ or $18K$ ground-truth programs. Thus, RNs can achieve very competitive, and even super-human results under much weaker and more natural assumptions, and even in situations when functional programs are unavailable.

\begin{table}
\centering
\small
\begin{threeparttable}
\begin{tabular}{l | C{14mm} | C{10mm} C{10mm} C{14mm} C{14mm} C{14mm}}
\toprule
Model & \textbf{Overall} & Count & Exist & Compare Numbers & Query Attribute & Compare Attribute \\
\hline
Human                 & 92.6 & 86.7 & 96.6 & 86.5 & 95.0 & 96.0 \\
\hline
Q-type baseline       & 41.8 & 34.6 & 50.2 & 51.0 & 36.0 & 51.3  \\
LSTM                  & 46.8 & 41.7 & 61.1 & 69.8 & 36.8 & 51.8  \\
CNN$+$LSTM            & 52.3 & 43.7 & 65.2 & 67.1 & 49.3 & 53.0  \\
CNN$+$LSTM$+$SA       & 68.5 & 52.2 & 71.1 & 73.5 & 85.3 & 52.3  \\
CNN$+$LSTM$+$SA*      & 76.6 & 64.4 & 82.7 & 77.4 & 82.6 & 75.4  \\
\hline
CNN+LSTM+RN & \textbf{95.5} & \textbf{90.1} & \textbf{97.8} & \textbf{93.6} & \textbf{97.9} & \textbf{97.1} \\
\bottomrule\addlinespace[1ex]
\end{tabular}
\begin{tablenotes}
\item * Our implementation, with optimized hyperparameters and trained fully end-to-end.
\end{tablenotes}
\end{threeparttable}
\caption{\textbf{Results on CLEVR from pixels.} Performances of our model (RN) and previously reported models \cite{johnson2017infer}, measured as accuracy on the test set and broken down by question category.}
\label{table:main}
\end{table}

\subsection{CLEVR from state descriptions}
\label{sec:CLEVRstate}
To demonstrate that the RN is robust to the form of its input, we trained our model on the state description matrix version of the CLEVR dataset. The model achieved an accuracy of $96.4\%$. This result demonstrates the generality of the RN module, showing its capacity to learn and reason about object relations while being agnostic to the kind of inputs it receives -- i.e., to the particular representation of the object features to which it has access. Therefore, RNs are not necessarily restricted to visual problems, and can thus be applied in very different contexts, and to different tasks that require relational reasoning.

\subsection{Sort-of-CLEVR from pixels}
\label{sec:results_sort_of_clevr}
The results so far led us to hypothesize that the difficulty in solving CLEVR lies in its heavy emphasis on relational reasoning, contrary to previous claims that the difficulty lies in question parsing \cite{kafle2017analysis}. However, the questions in the CLEVR dataset are not categorized based on the degree to which they may be relational, making it hard to assess our hypothesis. Therefore, we use the Sort-of-CLEVR dataset which we explicitly designed to seperate out relational and non-relational questions (see Section \ref{sec:sort_of_clevr}).

We find that a CNN augmented with an RN achieves an accuracy above $94\%$ for both relational and non-relational questions. However, a CNN augmented with an MLP only reached this performance on the non-relational questions, plateauing at $63\%$ on the relational questions. This strongly indicates that models lacking a dedicated relational reasoning component struggle, or may even be completely incapable of solving tasks that require very simple relational reasoning. Augmenting these models with a relational module, like the RN, is sufficient to overcome this hurdle. 

A simple ``closest-to'' or ``furthest-from'' relation is particularly revealing of a CNN+MLP's lack of general reasoning capabilities ($52.3\%$ success). For these relations a model must gauge the distances between \emph{each} object, and then compare each of these distances. Moreover, depending on the images, the relevant distance could be quite small in magnitude, or quite large, further increasing the combinatoric difficulty of this task.

\subsection{bAbI}
\label{sec:babi_results}
Our model succeeded on $18/20$ tasks. Notably, it succeeded on the basic induction task ($2.1\%$ total error), which proved difficult for the Sparse DNC ($54\%$), DNC ($55.1\%$), and EntNet ($52.1\%$). Also, our model did not catastrophically fail in any of the tasks: for the $2$ tasks that it failed (the ``two supporting facts'', and ``three supporting facts'' tasks), it missed the $95\%$ threshold by $3.1\%$ and $11.5\%$, respectively. We also note that the model we evaluated was chosen based on overall performance on a withheld validation set, using a single seed. That is, we did not run multiple replicas with the best hyperparameter settings (as was done in other models, such as the Sparse DNC, which demonstrated performance fluctuations with a standard deviation of more than $\pm3$ tasks passed for the best choice of hyperparameters). 

\subsection{Dynamic physical systems}
\label{sec:physical_systems}
Finally, we trained our model on two tasks requiring reasoning about the dynamics of balls moving along a surface. In the connection inference task, our model correctly classified all the connections in $93\%$ of the sample scenes in the test set. In the counting task, the RN achieved similar performance, reporting the correct number of connected systems for $95\%$ of the test scene samples. In comparison, an MLP with comparable number of parameters was unable to perform better than chance for both tasks. Moreover, using this task to learn to infer relations results in transfer to unseen motion capture data, where RNs predict the connections between body joints of a walking human (see supplementary information for experimental details and example videos).

\section{Discussion and Conclusions}
This work showed how the RN, a dedicated module for computing inter-entity relations, can be plugged into broader deep learning architectures to substantially improve performance on tasks that demand rich relational reasoning. Our CLEVR results included super-human performance at $95.5\%$ overall. Our bAbI results demonstrated broad reasoning capabilities, solving $18/20$ tasks with no catastrophic failures. Together these results demonstrate the flexibility and power of this simple neural network building block.

One of the most interesting aspects of the work is that RN module inclusion in relatively simple CNN- and LSTM-based VQA architectures raised the performance on CLEVR from $68.5\%$ to $95.5\%$ and achieved state-of-the-art, super-human performance. We speculate that the RN provided a more powerful mechanism for flexible relational reasoning, and freed up the CNN to focus more exclusively on processing local spatial structure. This distinction between \emph{processing} and \emph{reasoning} is important. Powerful deep learning architectures, such as ResNets, are highly capable visual processors, but they may not be the most appropriate choice for reasoning about arbitrary relations.

A key contribution of this work is that the RN was able to induce, through the learning process, upstream processing to provide a set of useful object-like representations. Note, the input data and target objective functions did not specify any particular form or semantics of the internal object representations. This demonstrates the RN's rich capacity for structured reasoning even with unstructured inputs and outputs.

Future work should apply RNs to a variety of problems that can benefit from structure learning and exploitation, such as rich scene understanding in RL agents, modeling social networks, and abstract problem solving. Future work could also improve the efficiency of RN computations. Though our results show that no knowledge about the particular relations among objects are necessary, RNs can exploit such knowledge if available or useful. For example, if two objects are known to have no actual relation, the RN's computation of their relation can be omitted. An important direction is exercising this option in circumstances with strict computational constraints, where, for instance, attentional mechanisms could be used to filter unimportant relations and thus bound the otherwise quadratic complexity of the number of considered pairwise relations. 

Relation Networks are a simple and powerful approach for learning to perform rich, structured reasoning in complex, real-world domains.

\vskip 0.5in
\subsection*{Acknowledgments}
We would like to thank Murray Shanahan, Ari Morcos, Scott Reed, Daan Wierstra, Alex Lerchner, and many others on the DeepMind team, for critical feedback and discussions.

\renewcommand{\thesubsection}{\Alph{subsection}}
\section*{Supplementary Material}
\label{sec:supplementary_material}


Here we provide additional details on (\ref{appendix:RelatedWork}) related work, (\ref{appendix:CLEVR}) CLEVR from pixels, (\ref{appendix:CLEVR_states}) CLEVR from state descriptions, (\ref{appendix:Sort_of_CLEVR}) Sort-of-CLEVR, (\ref{appendix:BABI}) bAbI, and (\ref{appendix:physical_systems}) Dynamic physical system reasoning. For each task, we provide additional information on the dataset, model architecture, training and results where necessary.

\subsection{Related Work}
\label{appendix:RelatedWork}
Since the RN is highly versatile, it can be used for visual, text-based, and state-based tasks. As such, it touches upon a broad range of areas in machine learning, computer vision, and natural language understanding. Here, we provide a brief overview of some of the most relevant related work.

\subsubsection*{Relational reasoning}
Relational reasoning is implicit in many symbolic approaches \cite{harnad1990symbol,newell1980physical} and has been explicitly pursued using neural networks as well \cite{das2016chains}. There is recent work applying neural networks to graphs, which are a natural structure for formalising relations \cite{henaff2015deep,kipf2016semi,niepert2016learning,scarselli2009graph,li2015gated,battaglia2016interaction}. Perhaps a crucial difference between this work and our work here is that RNs require minimal oversight to produce their input (a set of objects), and can be applied successfully to tasks even when provided with relatively unstructured inputs coming from CNNs and LSTMs. There has also been some recent work on reasoning about sets, although this work does not explicitly reason about the \emph{relations} of elements \emph{within} sets \cite{Zaheer2017DeepSets}.

\subsubsection*{Grounding spatial relations}
Although grounding language in spatial percepts has a long-standing tradition, the majority of previous research has focused on either rule-based spatial representations or hand-engineered spatial features \cite{golland2010game, guadarrama2013grounding, krishnamurthy2013jointly, kruijff2007situated, lan2012image, malinowski14nips, tellex2011understanding, tellex2010}. Although there are some attempts to learn spatial relations using spatial templates \cite{logan1996computational,malinowski2014pooling}, these approaches are less versatile than ours. 

\subsubsection*{Visual question answering}
Visual question answering is a recently introduced task that measures a machine understanding of the scene through questions \cite{antol2015vqa,malinowski14nips}. Related to our work, we are mostly interested in the newly introduced CLEVR dataset \cite{johnson2016clevr} that distills core challenges of the task, namely relational and multi-modal reasoning. The majority of approaches to question answering share the same pipeline \cite{gao2015you,malinowski2016ask, ren2015image}. First, questions are encoded with recurrent neural networks, and images are encoded with convolutional neural networks. Next, both representations are combined, and the answers are either predicted or generated. Most successful methods also use an attention mechanism that locate important image regions \cite{fukui2016multimodal, xu2016ask, xu2015show, yang2016stacked}. In our work, we follow a similar pipeline, but we use Relation Networks as a powerful reasoning module.

Parallel to our work, two architectures have shown impressive results on the CLEVR dataset \cite{hu2017learning, johnson2017infer}. Both approaches hinge on compositionality principles, and have shown they are capable of some relational reasoning. However, both require either designing modules, or require direct access to ground-truth programs. The RN module, on the other hand, is conceptually simpler, can readily be combined with basic neural components such as CNNs or LSTMs, can be broadly  applied to various tasks, and achieves significantly better results on CLEVR \cite{johnson2016clevr} than \cite{hu2017learning}, and on par with strongly supervised system of \cite{johnson2017infer}. 

\subsubsection*{Text-based question answering}
Answering text-based questions has long been an active research area in the NLP community \cite{berant2013semantic, kwiatkowski2010inducing, liang2013learning, zelle1996learning}. Recently, in addition to traditional symbolic-based question answering architectures, we observe a growing interest in neural-based approaches to text based question answering \cite{rae2016scaling,weston2014memory,xiong2016dynamic}. While these architectures rely on `memories', we empirically show that the RN module has similar capabilities, reaching very competitive results on the bAbI dataset \cite{weston2015towards} -- a dataset that test reasoning capabilities of text-based question answering models.

\subsection{CLEVR from pixels}
\label{appendix:CLEVR}

Our model (described in Section 4 of the main text) was trained on $70000$ scenes from the CLEVR dataset and a total of $699989$ questions. Images were first down-sampled to size $128\times128$, then pre-processed with padding to size $136\times136$, followed by random cropping back to size $128\times128$ and slight random rotations between $-0.05$ and $0.05$ rads. We used $10$ distributed workers that synchronously updated a central parameter server. Each worker learned with mini-batches of size $64$, using the Adam optimizer and a learning rate of $2.5e^{-4}$. Dropout of $50\%$ was used on the penultimate layer of the RN. In our best performing model each convolutional layer used 24 kernels of size $3\times3$ and stride $2$, batch normalization, and rectified linear units. The model stopped improving in performance after approximately $1.4$ million iterations, at which point training was concluded. The model achieved $96.8\%$ accuracy on the validation set. In general, we found that smaller models performed best. For example, $128$ hidden unit LSTMs performed better than $256$ or $512$, and CNNs with $24$ kernels were better than CNNs with more kernels, such as $32$, $64$, or more.

\subsubsection*{Failure cases} 
Although our model gets most answers correct, a closer examination of the failure cases help us to identify limitations of our architecture. In Table \ref{fig:clevr_failures}, we show some examples of CLEVR questions that our model fails to answer correctly, along with the ground-truth answers. Based on our observations, we hypothesize that our architecture fails especially when objects are heavily occluded, or whenever a high precision object position representation is required. We also observe that many failure cases for our model are also challenging for humans.

\begin{table*}[h!]
\footnotesize
\centering
\begin{center}
\begin{tabular}{l@{\ }c@{\ }c@{\ }c}
\multicolumn{2}{c}{\includegraphics[width=0.27\linewidth]{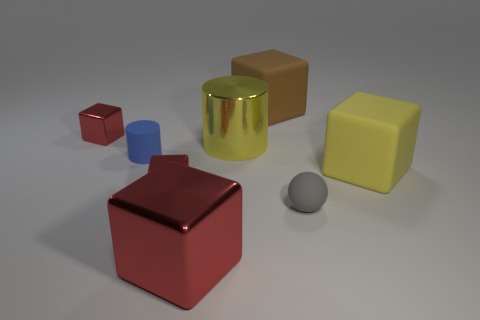}} &
\includegraphics[width=0.27\linewidth]{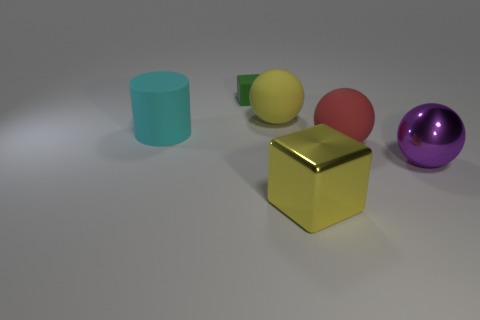} &
\includegraphics[width=0.27\linewidth]{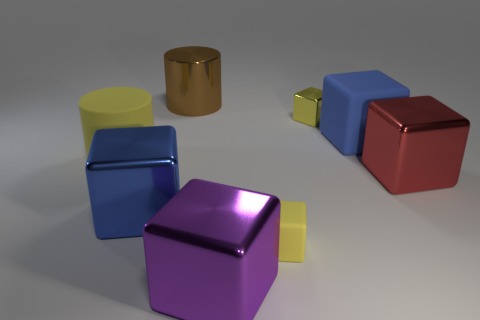} 
\\
\multicolumn{2}{c}{\scriptsize What shape is the small object} & \multicolumn{1}{c}{\scriptsize What number of things are either tiny} & 
\multicolumn{1}{c}{\scriptsize What number of objects are blocks}
\\
\multicolumn{2}{c}{\scriptsize that is in front of the yellow matte} & \multicolumn{1}{c}{\scriptsize green rubber objects or shiny things} & 
\multicolumn{1}{c}{\scriptsize that are in front of the large}
\\
\multicolumn{2}{c}{\scriptsize thing and behind the gray sphere?} & \multicolumn{1}{c}{\scriptsize that are behind the big metal block?} & 
\multicolumn{1}{c}{\scriptsize red cube or green balls?}
\\\midrule
\textit{\footnotesize RN:}&\multicolumn{1}{c}{\footnotesize\textcolor{red}{cylinder}} & \multicolumn{1}{c}{\footnotesize\textcolor{red}{1}} & \multicolumn{1}{c}{\footnotesize\textcolor{red}{2}}
\\\midrule
 \textit{\footnotesize GT:} &\multicolumn{1}{c}{\footnotesize\textcolor{green}{cube}} & 
\multicolumn{1}{c}{\footnotesize\textcolor{green}{2}} & 
\multicolumn{1}{c}{\footnotesize\textcolor{green}{3}}\\
\midrule

\multicolumn{2}{c}{\includegraphics[width=0.27\linewidth]{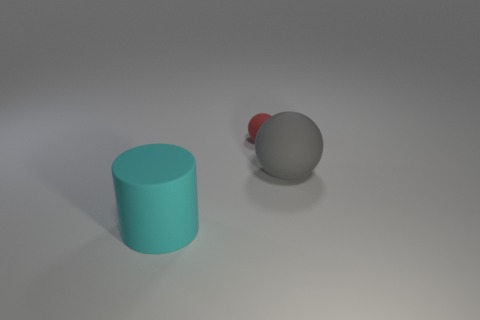}} &
\includegraphics[width=0.27\linewidth]{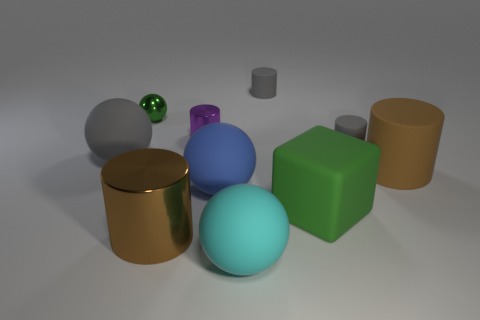} &
\includegraphics[width=0.27\linewidth]{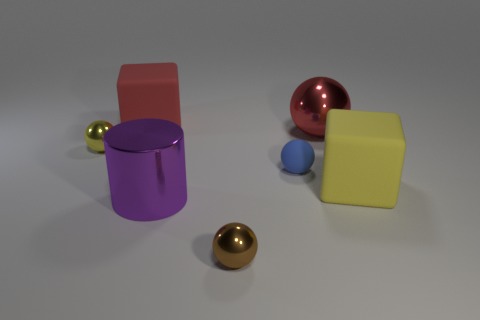} 
\\
\multicolumn{2}{c}{\scriptsize Is the shape of the small red object} & \multicolumn{1}{c}{\scriptsize How many gray objects are in front} & 
\multicolumn{1}{c}{\scriptsize What number of objects are big}
\\
\multicolumn{2}{c}{\scriptsize  the same as the large matte object} & \multicolumn{1}{c}{\scriptsize of the tiny green shiny ball and right} & 
\multicolumn{1}{c}{\scriptsize  red matte cubes or things on the right}
\\
\multicolumn{2}{c}{\scriptsize  that is right of the small rubber ball?} & \multicolumn{1}{c}{\scriptsize   of the big blue matte thing?} & 
\multicolumn{1}{c}{\scriptsize side of the large red matte block?}
\\\midrule
\textit{RN:}&\multicolumn{1}{c}{\footnotesize\textcolor{red}{no}} & \multicolumn{1}{c}{\footnotesize\textcolor{red}{0}} & \multicolumn{1}{c}{\footnotesize\textcolor{red}{5}}
\\\midrule
 \textit{GT:} &\multicolumn{1}{c}{\footnotesize\textcolor{green}{yes}} & 
\multicolumn{1}{c}{\footnotesize\textcolor{green}{1}} & 
\multicolumn{1}{c}{\footnotesize\textcolor{green}{6}}\\
\midrule

\multicolumn{2}{c}{\includegraphics[width=0.27\linewidth]{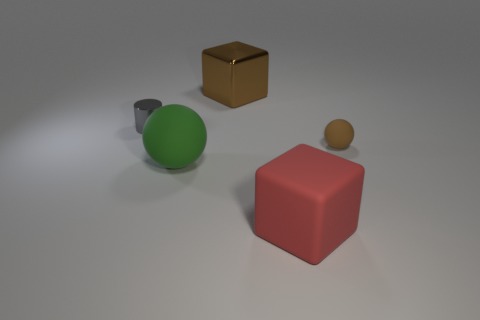}} &
\includegraphics[width=0.27\linewidth]{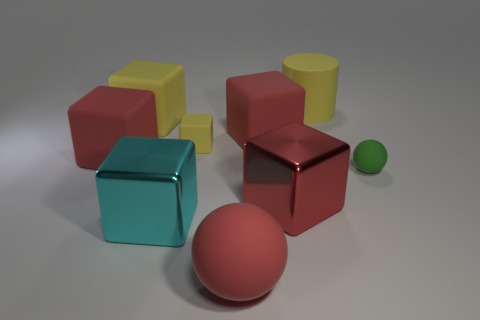} &
\includegraphics[width=0.27\linewidth]{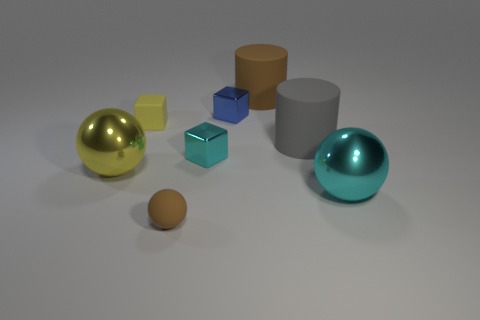} 
\\
\multicolumn{2}{c}{\scriptsize There is a brown ball;} & \multicolumn{1}{c}{\scriptsize  How many objects are big purple} & 
\multicolumn{1}{c}{\scriptsize How many things are rubber}
\\
\multicolumn{2}{c}{\scriptsize  what number of things are left of it?} & \multicolumn{1}{c}{\scriptsize  rubber blocks or red blocks in front} & 
\multicolumn{1}{c}{\scriptsize   cylinders in front of the tiny yellow}
\\
\multicolumn{2}{c}{\scriptsize } & \multicolumn{1}{c}{\scriptsize    of the tiny yellow rubber thing?} & 
\multicolumn{1}{c}{\scriptsize   block or blocks that are to the right }
\\
\multicolumn{2}{c}{\scriptsize } & \multicolumn{1}{c}{\scriptsize} & 
\multicolumn{1}{c}{\scriptsize  of the small brown rubber thing?}
\\\midrule
\textit{RN:}&\multicolumn{1}{c}{\footnotesize\textcolor{red}{3}} & \multicolumn{1}{c}{\footnotesize\textcolor{red}{3}} & \multicolumn{1}{c}{\footnotesize\textcolor{red}{2}}
\\\midrule
 \textit{GT:} &\multicolumn{1}{c}{\footnotesize\textcolor{green}{4}} & 
\multicolumn{1}{c}{\footnotesize\textcolor{green}{2}} & 
\multicolumn{1}{c}{\footnotesize\textcolor{green}{3}}\\
\midrule

\multicolumn{2}{c}{\includegraphics[width=0.27\linewidth]{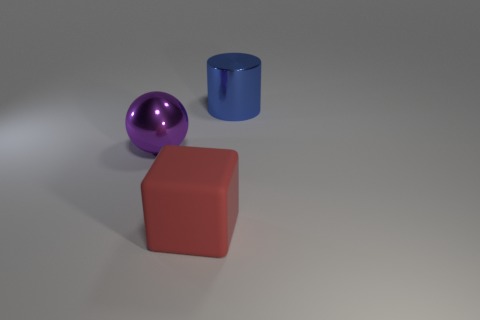}} &
\includegraphics[width=0.27\linewidth]{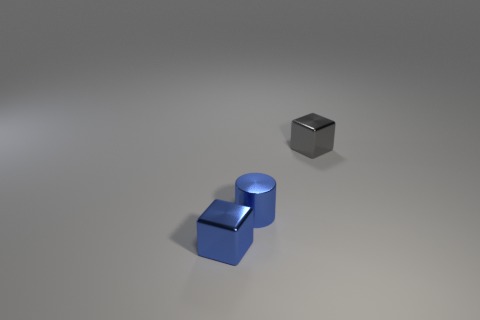} &
\includegraphics[width=0.27\linewidth]{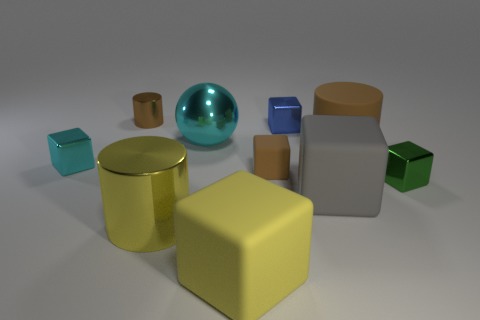} 
\\
\multicolumn{2}{c}{\scriptsize  What number of objects are either } & \multicolumn{1}{c}{\scriptsize  Are there the same number of small} & 
\multicolumn{1}{c}{\scriptsize What number of other things}
\\
\multicolumn{2}{c}{\scriptsize  big things that are left} & \multicolumn{1}{c}{\scriptsize   blue objects that are to the right of} & 
\multicolumn{1}{c}{\scriptsize  are there of the same}
\\
\multicolumn{2}{c}{\scriptsize  of the cylinder or cylinders?} & \multicolumn{1}{c}{\scriptsize  the blue cube and blue metal cubes?} & 
\multicolumn{1}{c}{\scriptsize  material as the green cube?}
\\\midrule
\textit{RN:}&\multicolumn{1}{c}{\footnotesize\textcolor{red}{2}} & \multicolumn{1}{c}{\footnotesize\textcolor{red}{no}} & \multicolumn{1}{c}{\footnotesize\textcolor{red}{6}}
\\\midrule
 \textit{GT:} &\multicolumn{1}{c}{\footnotesize\textcolor{green}{3}} & 
\multicolumn{1}{c}{\footnotesize\textcolor{green}{yes}} & 
\multicolumn{1}{c}{\footnotesize\textcolor{green}{5}}\\
\midrule

\end{tabular}
\end{center}
\caption{Failures on CLEVR; RN -- predicted answers, GT -- ground-truth answer. 
}
\label{fig:clevr_failures}
\end{table*}

\subsection{CLEVR from state descriptions} 
\label{appendix:CLEVR_states}

The model that we train on the state description version of CLEVR is similar to the model trained on the pixel version of CLEVR, but without the vision processing module. We used a $256$ unit LSTM for question processing and word-lookup embeddings of size $32$. For the RN we used a four-layer MLP with $512$ units per layer, with ReLU non-linearities for $g_\theta$. A three-layer MLP consisting of $512$, $1024$ (with $2\%$ dropout) and $29$ units with ReLU non-linearities was used for $f_\theta$. To train the model we used $10$ distributed workers that synchronously updated a central parameter server. Each worker learned with mini-batches of size $64$, using the Adam optimizer and a learning rate of $1e^{-4}$.

\subsection{Sort-of-CLEVR} 
\label{appendix:Sort_of_CLEVR}

The Sort-of-CLEVR dataset contains $10000$ images of size $75\times75$, $200$ of which were withheld for validation. There were $20$ questions generated per image ($10$ relational and $10$ non-relational).

Non-relational questions are split into three categories: (i) query shape, e.g. ``\textit{What is the shape of the red object?}''; (ii) query horizontal position, e.g. ``\textit{Is the red object on the left or right of the image?}''; (iii) query vertical position, e.g. ``\textit{Is the red object on the top or bottom of the image?}''. These questions are non-relational because one can answer them by reasoning about the attributes  (e.g. position, shape) of a single entity which is identified by its unique color (e.g. \textit{red}).

Relational questions are split into three categories: (i) closest-to, e.g. ``\textit{What is the shape of the object that is closest to the green object?}''; (ii) furthest-from, e.g. ``\textit{What is the shape of the object that is furthest from the green object?}''; (iii) count, e.g. ``\textit{How many objects have the shape of the green object?}''. We consider these relational because answering them requires reasoning about the attributes of one or more objects that are defined relative to the attributes of a reference object. This reference object is uniquely identified by its color.

Questions were encoded as binary strings of length $11$, where the first $6$ bits identified the color of the object to which the question referred, as a one-hot vector, and the last $5$ bits identified the question type and subtype.

In this task our model used: four convolutional layers with $32$, $64$, $128$ and $256$ kernels, ReLU non-linearities, and batch normalization; the questions, which were encoded as fixed-length binary strings, were treated as question embeddings and passed directly to the RN alongside the object pairs; a four-layer MLP consisting of $2000$ units per layer with ReLU non-linearities was used for $g_{\theta}$; and a four-layer MLP consisting of $2000$, $1000$, $500$, and $100$ units with ReLU non-linearities used for $f_{\phi}$. An additional final linear layer produced logits for a softmax over the possible answers. The softmax output was optimized with a cross-entropy loss function using the Adam optimizer with a learning rate of $1e^{-4}$ and mini-batches of size $64$.

We also trained a comparable MLP based model (CNN+MLP model) on the Sort-of-CLEVR task, to explore the extent to which a standard model can learn to answer relational questions. We used the same CNN and LSTM, trained end-to-end, as described above. However, this time we replaced the RN with an MLP with the same number of layers and number of units per layer. Note that there are more parameters in this model because the input layer of the MLP connects to the full CNN image embedding.

\begin{figure}[h!]
    \centering
    \includegraphics[width=0.95\textwidth]{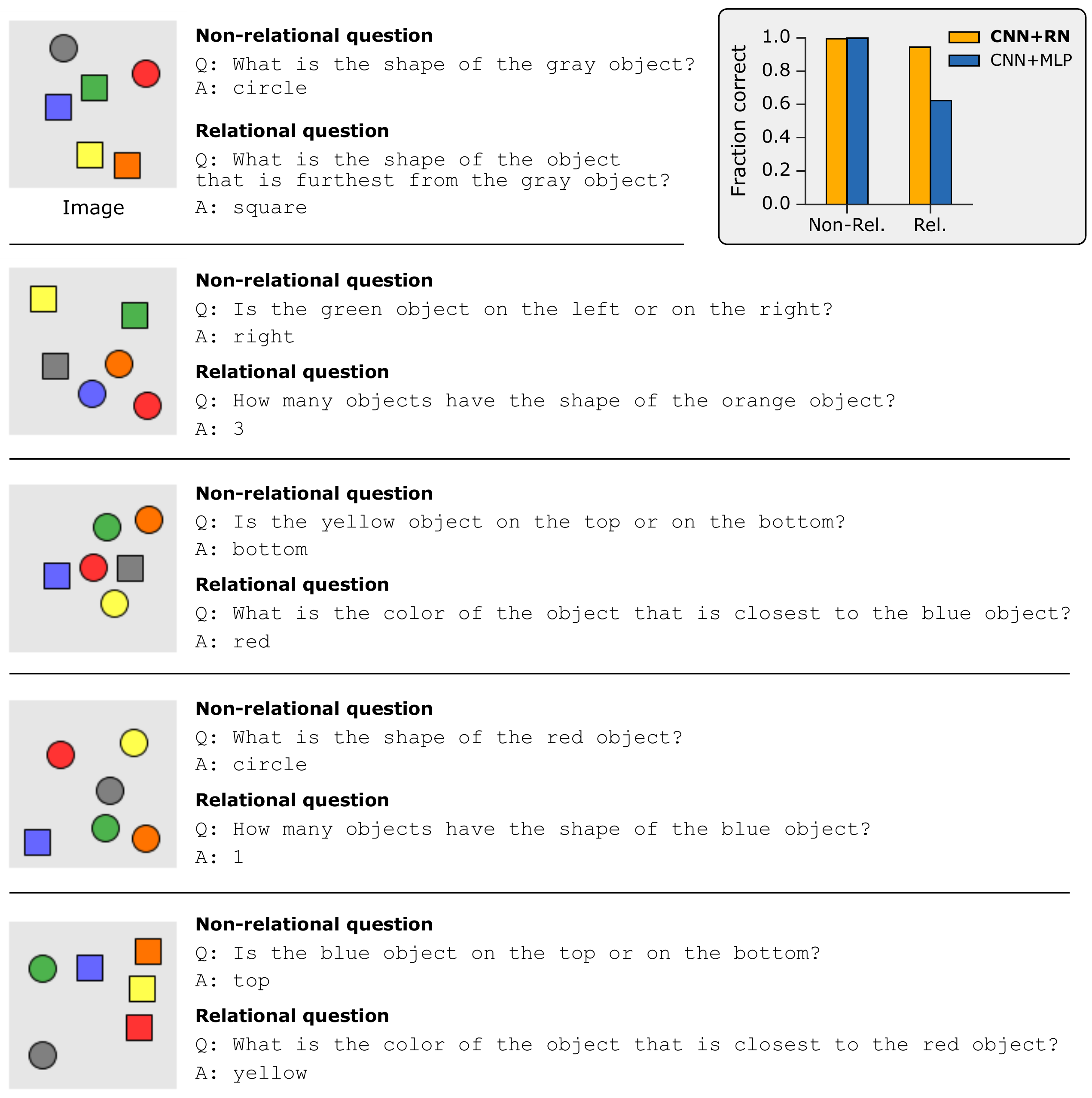}
    \caption{\textbf{``Sort-of-CLEVR'' task: examples and results.} The Sort-of-CLEVR example here consists of an image of six objects and two questions -- a relational question, and a non-relational question -- along with the corresponding answers. The fraction of correctly answered relational questions (inset bar plot) for our model (CNN+RN) is much larger than the comparable MLP based model (CNN+MLP), whereas both models have similar performance levels for non-relational questions.}
    \label{fig:soc_example}
\end{figure}

\subsection{bAbI model for language understanding} 
\label{appendix:BABI}

For the bAbI task, each of the $20$ sentences in the support set was processed through a $32$ unit LSTM to produce an object. For the RN, $g_{\theta}$ was a four-layer MLP consisting of $256$ units per layer. For $f_{\phi}$, we used a three-layer MLP consisting of $256$, $512$, and $159$ units, where the final layer was a linear layer that produced logits for a softmax over the answer vocabulary. A separate LSTM with $32$ units was used to process the question. The softmax output was optimized with a cross-entropy loss function using the Adam optimizer with a learning rate of $2e^{-4}$.

\subsection{Dynamic physical system reasoning} 
\label{appendix:physical_systems}

For the connection inference task the targets were binary vectors representing the existence (or non-existence) of a connection between each ball pair. For a total of $10$ objects, the targets were $10^2$ length vectors. For the counting task, the targets were one-hot vectors (of length $10$) indicating the number of systems of connected balls. It is important to point out that in the first task the supervision signal provided by the targets explicitly informs about the relations that need to be computed. In the second task, the supervision signal (counts of systems) do not provide explicit information about the kind of relations that need to be computed. Therefore, the models that solve the counting task must successfully infer the relations implicitly.

Inputs to the RN were state descriptions. Each row of a state description matrix provided information about a particular object (i.e. ball), including its coordinate position and color. Since the system was dynamic, and hence evolved through time, each row contained object property descriptions for $16$ consecutive time-frames. For example, a row could be comprised of $33$ floats: $16$ for the object's $x$ coordinate position across $16$ frames, $16$ for the object's $y$ coordinate position across $16$ frames, and $1$ for the object's color. The RN treated each row in this state description matrix as an object. Thus, it had to infer an object description contained information of the object's properties evolving through time.

For the connection inference task, the RN's $g_{\theta}$ was a four-layer MLP consisting of three layers with $1000$ units and one layer with $500$ units. For $f_{\phi}$, we used a three-layer MLP consisting of $500$, $100$, and $100$ units, where the final layer was a linear layer that produced logits corresponding to the existence/absence of a connection between each ball pair. The output was optimized with a cross-entropy loss function using the Adam optimizer with a learning rate of $1e^{-4}$ and a batch size of $50$. The same model was used for the counting task, but this time the output layer of the RN was a linear layer with $10$ units. For baseline comparisons we replaced the RNs with MLPs with comparable number of parameters.

Please see the supplementary videos: 

\url{https://www.youtube.com/channel/UCIAnkrNn45D0MeYwtVpmbUQ}

\clearpage
\small
\bibliography{references}

\end{document}